\documentclass[runningheads]{llncs}
\usepackage{cite}
\usepackage{graphicx}
\usepackage{epstopdf}
\usepackage{array}
\usepackage{amsmath,amssymb} 
\usepackage{color}
\usepackage{url}
\usepackage{subfig}
\usepackage{overpic}
\usepackage{bm}

\begin{document}
\title{Bi-GANs-ST for Perceptual Image Super-resolution}

\titlerunning{Bi-GANs-ST for Perceptual Image Super-resolution}
%
\author{Xiaotong Luo\inst{1} \and
Rong Chen\inst{1} \and
Yuan Xie\inst{2} \and
Yanyun Qu\thanks{Corresponding author}\inst{1} \and
Cuihua Li\inst{1}}
%
\authorrunning{X. Luo, R. Chen, Y. Xie, Y. Qu and C. Li}
%

\institute{School of Information Science and Engineering, Xiamen University, Xiamen, China
\email{528524296@qq.com, chenrong\_mail@qq.com, \{yyqu,chli\}@xmu.edu.cn}\\ \and
Research Center of Precision Sensing and Control, Institute of Automation, Chinese Academy of Sciences, Beijing, China\\
\email{yuan.xie@ia.ac.cn}}
\maketitle              
\begin{abstract}
Image quality measurement is a critical problem for image super-resolution (SR) algorithms. Usually, they are evaluated by some well-known objective metrics, \emph{e.g.}, PSNR and SSIM, but these indices cannot provide suitable results in accordance with the perception of human being. Recently, a more reasonable perception measurement has been proposed in \cite{blau2017perception:9}, which is also adopted by the PIRM-SR $2018$ challenge. In this paper, motivated by \cite{blau2017perception:9}, we aim to generate a high-quality SR result which balances between the two indices, {\it i.e.,} the perception index and root-mean-square error (RMSE). To do so, we design a new deep SR framework, dubbed Bi-GANs-ST, by integrating two complementary generative adversarial networks (GAN) branches. One is memory residual SRGAN (MR-SRGAN), which emphasizes on improving the {\it objective} performance, such as reducing the RMSE. The other is weight perception SRGAN (WP-SRGAN), which obtains the result that favors better {\it subjective} perception via a two-stage adversarial training mechanism. Then, to produce final result with excellent perception scores and RMSE, we use soft-thresholding method to merge the results generated by the two GANs. Our method performs well on the perceptual image super-resolution task of the PIRM $2018$ challenge. Experimental results on five benchmarks show that our proposal achieves highly competent performance compared with other state-of-the-art methods.

\end{abstract}
\section{Introduction}

Single image super-resolution (SISR) is a hotspot in image restoration. It is an inverse problem which recovers a high-resolution (HR) image from a low-resolution (LR) image via super-resolution (SR) algorithms. Traditional SR algorithms are inferior to deep learning based SR algorithms on speed and some distortion measures, \emph{e.g.}, peak signal-to-noise ratio (PSNR) and structural similarity index (SSIM). In addition, SR algorithms based on deep learning can also obtain excellent visual effects \cite{dong2016image:1,kim2016accurate:2,dong2016accelerating:3,ledig2017photo:4,lai2017deep:5,lim2017enhanced:6,haris2018deep:7}.

Here, SR algorithms with deep learning can be divided into two categories. One is built upon convolutional neural network with  classic L1 or L2 loss in pixel space as the optimization function, which can gain a higher PSNR but over-smoothness for lacking enough high-frequency texture information. The representative approaches are SRResNet \cite{ledig2017photo:4} and EDSR \cite{lim2017enhanced:6}. The other is based on generative adversarial networks (GAN), {\it e.g.,} SRGAN \cite{ledig2017photo:4} and EnhanceNet \cite{sajjadi2017enhancenet:8}, which introduces perceptual loss in the optimization function. This kind of algorithms can restore more details and improve visual performance at the expense of objective evaluation indices. Different quality assessment methods are used in various application scenarios. For example, medical imaging may concentrate on objective evaluation metrics, while the subjective visual perception may be more important for natural images. Therefore, we need to make a balance between the objective evaluation criteria and subjective visual effects.

Blau \emph{et al.} \cite{blau2017perception:9} proposed perceptual-distortion plane which jointly quantified the accuracy and perceptual quality of algorithms and also pointed GAN can make the perceptual-distortion tradeoff. In the PIRM-SR $2018$ challenge \cite{Blau2018real:12}, image quality is evaluated by root-mean-square error (RMSE) and perceptual index. Inspired by \cite{blau2017perception:9}, we design a new SR framework for perceptual image super-resolution which includes two GAN branches. First, we redesign the generator network based on SRGAN in each branch and adopt two-stage adversarial training mechanism in the second branch. Then, soft-thresholding method is used to fuse the two results generated by the two branches. Experimental results show our method can obtain excellent distortion measurement and perceptual quality. The contributions of our algorithm are three-fold:

1) We propose a new SR framework named Bi-branch GANs with Soft-thresholding (Bi-GANs-ST) for perceptual image super-resolution which consists of two branches. The one is memory residual SRGAN (MR-SRGAN) which emphases on improving the objective performance (\emph{e.g.}, reduce the RMSE value). The other is weight perception SRGAN (WP-SRGAN) which focuses on better subjective perception (\emph{e.g.}, reduce the perceptual index).

2) In MR-SRGAN, we add memory storage mechanism in Generator which can improve the feature selection ability of the model. To further reduce the RMSE, we train MR-SRGAN by removing the logarithm of adversarial losses. In WP-SRGAN, we use two-stage adversarial training mechanism in which we first optimize pixel-wise loss as a pre-training model for obtaining lower RMSE, then optimize perceptual loss for reducing the perceptual index. And we remove Batch Normalization layers in both networks.

3) To keep balance between the perceptual index and RMSE, we fuse the results generated by MR-SRGAN and WP-SRGAN via soft-thresholding method. Our proposal achieves competent performance on the task of the PIRM-SR 2018 challenge.

\section{Related Work}

Abundant single image super-resolution algorithms based on deep learning have been proposed and achieved remarkable performance. Here, we mainly discuss image SR using deep neural networks, image SR using generative adversarial networks and image quality evaluation.

\subsection{Image super-resolution using deep neural networks}

Dong \emph{et al.} proposed SRCNN \cite{dong2016image:1}, which is a preliminary work to apply convolutional neural network into SISR. Although the network contains only three layers, the performance has been greatly improved compared with the traditional reconstructed methods. FSRCNN \cite{dong2016accelerating:3} is an accelerated version of SRCNN, which introduced a deconvolution layer at the end of the network to perform upsampling for reducing the computational complexity. Shi \emph{et al.} proposed ESPCN \cite{shi2016real:13}, which mainly utilized the sub-pixel convolutional layer to accelerate the training process. Kim \emph{et al.} proposed VDSR \cite{kim2016accurate:2}, which used cascaded filters and residual learning to obtain a larger receptive field and accelerate convergence. Kim \emph{et al.} \cite{kim2016deeply:14} first applied the recursive neural network and skip connection \cite{he2016deep:15} to image SR. RED network \cite{mao2016image:16} was composed of symmetric convolutional layers and deconvolution layers to learn the end-to-end mapping from LR to HR image pairs. Lai \emph{et al.} \cite{lai2017deep:5} proposed a cascaded pyramid structure with two branches, one is for feature extraction, the other is for image reconstruction. Moreover, Charbonnier loss was applied to multiple levels and it can generate sub-band residual images at each level. Tong \emph{et al.} \cite{tong2017image:17} introduced dense blocks combining low-level features and high-level features to improve the performance effectively. Lim \emph{et al.} \cite{lim2017enhanced:6} removed Batch Normalization layers in residual blocks (ResBlocks) and adopted residual scaling factor to stabilize network training. Besides, it also proposed multi-scale SR algorithm via a single network. However, when the scaling factor is equal to or larger than $4\times$, the results obtained by the aforementioned methods mostly look smooth and lack enough high-frequency details. The reason is that the optimization targets are mostly based on minimizing L1 or L2 loss in pixel space without considering the high-level features.

\subsection{Image super-resolution using generative adversarial networks}

{\bfseries Super-resolution with adversarial training.} Generative adversarial nets (GANs) \cite{goodfellow2014generative:18} consist of Generator and Discriminator. In the task of super-resolution, \emph{e.g.}, SRGAN \cite{ledig2017photo:4},  Generator is used to generate SR images. Discriminator distinguishes whether an image is true or forged. The goal of Generator is to generate a realistic image as much as possible to fool Discriminator. And Discriminator aims to distinguish the ground truth from the generated SR image. Thus, Generator and Discriminator constitute an adversarial game. With adversarial training, the forged data and the real data can eventually obey a similar image statistics distribution. Therefore, adversarial learning in SR is important for recovering the image textural statistics.

{\noindent\bfseries Perceptual loss for deep learning.} In order to be better accordant with human perception, Johnson \emph{et al.} \cite{johnson2016perceptual:19} introduced perceptual loss based on high-level features extracted from pre-trained networks, \emph{e.g.} VGG16, VGG19, for the task of style transfer and SR. Ledig \emph{et al.} \cite{ledig2017photo:4} proposed SRGAN, which aimed to make the SR images and the ground-truth (GT) similar not only in low-level pixels, but also in high-level features. Therefore, SRGAN can generate realistic images. Sajjadi \emph{et al.} proposed EnhanceNet \cite{sajjadi2017enhancenet:8}, which applied a similar approach and introduced the local texture matching loss, reducing visually unpleasant artifacts. Zhang \emph{et al.} \cite{zhang2018unreasonable:25} explained why the perceptual loss based on deep features fits human visual perception well. Mechrez \emph{et al.} proposed contextual loss \cite{mechrez2018contextual:20,mechrez2018learning:21} which was based on the idea of natural image statistics, and it is the best algorithm for recovering perceptual results in previous published works currently. Although these algorithms can obtain better perceptual image quality and visual performance, it cannot achieve better results in terms of objective evaluation criteria.

\subsection{Image quality evaluation}
 There are two ways to evaluate image quality including objective and subjective assessment criteria. The popular objective criteria includes the following: PSNR, SSIM, multi-scale structure similarity index (MSSSIM), information fidelity criterion (IFC), weighted peak signal-to-noise ratio (WPSNR), noise quality measure (NQM) \cite{yang2014single:22} and so on. Although IFC has the highest correlation with perceptual scores for SR evaluation \cite{yang2014single:22}, it is not the best criterion to assess the image quality. The subjective assessment is usually scored by human subjects in the previous works \cite{moorthy2011blind:23,mittal2012no:24}. However, there is not a suitable objective evaluation in accordance with the human subjective perception yet. In the PIRM-SR $2018$ challenge \cite{Blau2018real:12}, the assessment of perceptual image quality is proposed which combines the quality measures of Ma \cite{ma2017learning:10} and NIQE \cite{mittal2013making:11}. The formula of perceptual index is represented as follows,
\begin{align}
  Perceptual\ index=\frac{1}{2}((10-Ma)+NIQE)\label{eq1}
\end{align}
Here, a lower perceptual index indicates better perceptual quality.
\section{Proposed Methods}
We first describe the overall structure of Bi-GANs-ST and then construct the networks MR-SRGAN and WP-SRGAN. The soft thresholding method is used for image fusion, as presented in Section 3.4.

\subsection{Basic architecture of Bi-GANs-ST}
\begin{figure}
\centering
\setlength{\abovecaptionskip}{0.cm}
\setlength{\belowcaptionskip}{-0.cm}
\includegraphics[scale=0.6]{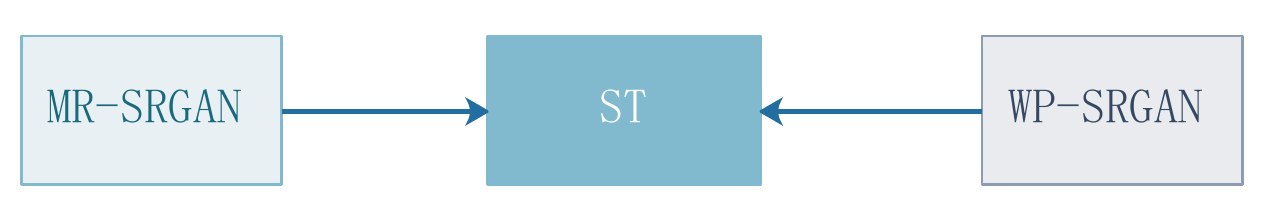}\\
 \begin{overpic}[scale=0.5]{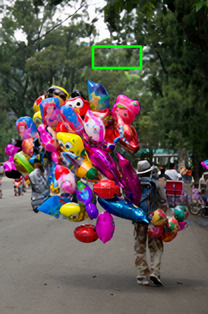} \put(0,0){\includegraphics[scale=0.68]{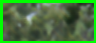}} \end{overpic} \
 \begin{overpic}[scale=0.5]{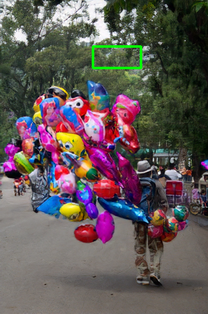} \put(0,0){\includegraphics[scale=0.68]{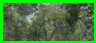}} \end{overpic} \
 \begin{overpic}[scale=0.5]{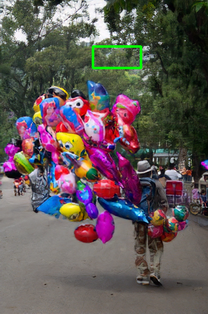} \put(0,0){\includegraphics[scale=0.68]{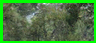}} \end{overpic} \vspace{-0mm}
\caption{The architecture of Bi-GANs-ST.}
\label{fig:1}
\end{figure}

As shown in Fig. \ref{fig:1}, our Bi-GANs-ST mainly consists of three parts: 1) memory residual SRGAN (MR-SRGAN), 2) weight perception SRGAN (WP-SRGAN), 3) soft thresholding (ST). The two GANs are used for generating two complementary SR images, and ST fuses the two SR results for balancing the perceptual score and RMSE.
\subsection{MR-SRGAN}
\begin{figure}
\centering
\includegraphics[scale=0.45]{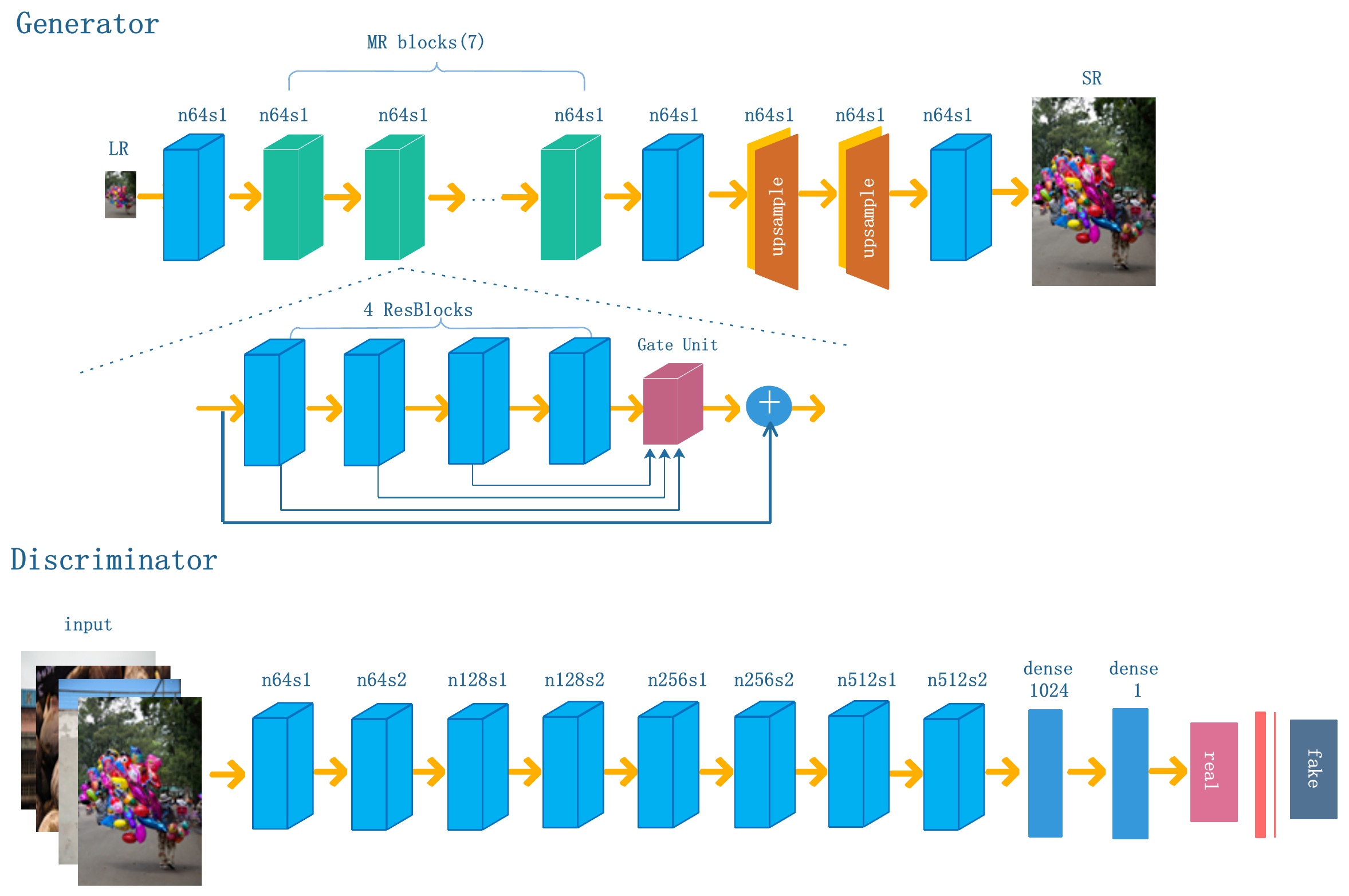}
\caption{The architecture of MR-SRGAN.}
\label{fig:2}
\end{figure}

{\bfseries Network architecture.} As illustrated in Fig. \ref{fig:2}, our MR-SRGAN is composed of Generator and Discriminator. In Generator, LR images are input to the network followed by one Conv layer for extracting shallow features. Then four memory residual (MR) blocks are applied for improving image quality which help to form persistent memory and improve the feature selection ability of model like MemEDSR \cite{chen2018persistent:26}. Each MR block consists of four ResBlocks and a gate unit. The former generates four-group features and then we extract a certain amount of features from these features by the gate unit. And the input features are added to the extracted features as the output of MR block. In ResBlocks, all the activation function layers are replaced with parametric rectified linear unit (PReLU) function and all the Batch Normalization (BN) layers are discarded in the generator network for reducing computational complexity. Finally, we restore the original image size by two upsampling operations. n is the corresponding number of feature maps and s denotes the stride for each convolutional layer in Fig. \ref{fig:2} and Fig. \ref{fig:3}. In Discriminator, we use the same setting as SRGAN \cite{ledig2017photo:4}.

{\noindent\bfseries Loss function.} The total generator loss function can be represented as three parts: pixel-wise loss, adversarial loss and perceptual loss, the formulas are as follows,
\begin{align}
 L_{total}=L_{pixel}+\lambda_1 L_{adv}+\lambda_2 L_{vgg}\label{eq2}
\end{align}
\begin{align}
 L_{pixel}=\frac{1}{N}\sum_{i=1}^{N}\left| x_{t}^{i}-G(x_{l}^{i}) \right |^{2}\label{eq3}
\end{align}
\begin{align}
 L_{adv}=-(D(G(x_{l})))\label{eq4}
\end{align}
\begin{align}
 L_{vgg}=\frac{1}{N}\sum_{i=1}^{N}\left | \phi (x_{t}^{i}) - \phi (G(x_{l}^{i}))\right |^{2}\label{eq5}
\end{align}
where $L_{pixel}$ is the pixel-wise MSE loss between the generated images and the ground truth, $L_{vgg}$ is the perceptual loss which calculates MSE loss between features extracted from the pre-trained VGG16 network, and $L_{adv}$ is the adversarial loss for Generator in which we remove logarithm. $\lambda_1$, $\lambda_2$ are the weights of adversarial loss and perceptual loss. $x_{t}$, $x_{l}$ denote the ground truth and LR images, respectively. $G(x_{l})$ is the SR images forged by Generator. $N$ represents the number of training samples. $\phi$ represents the features extracted from pre-trained VGG16 network.

\subsection{WP-SRGAN}
\begin{figure}
\centering
\includegraphics[scale=0.45]{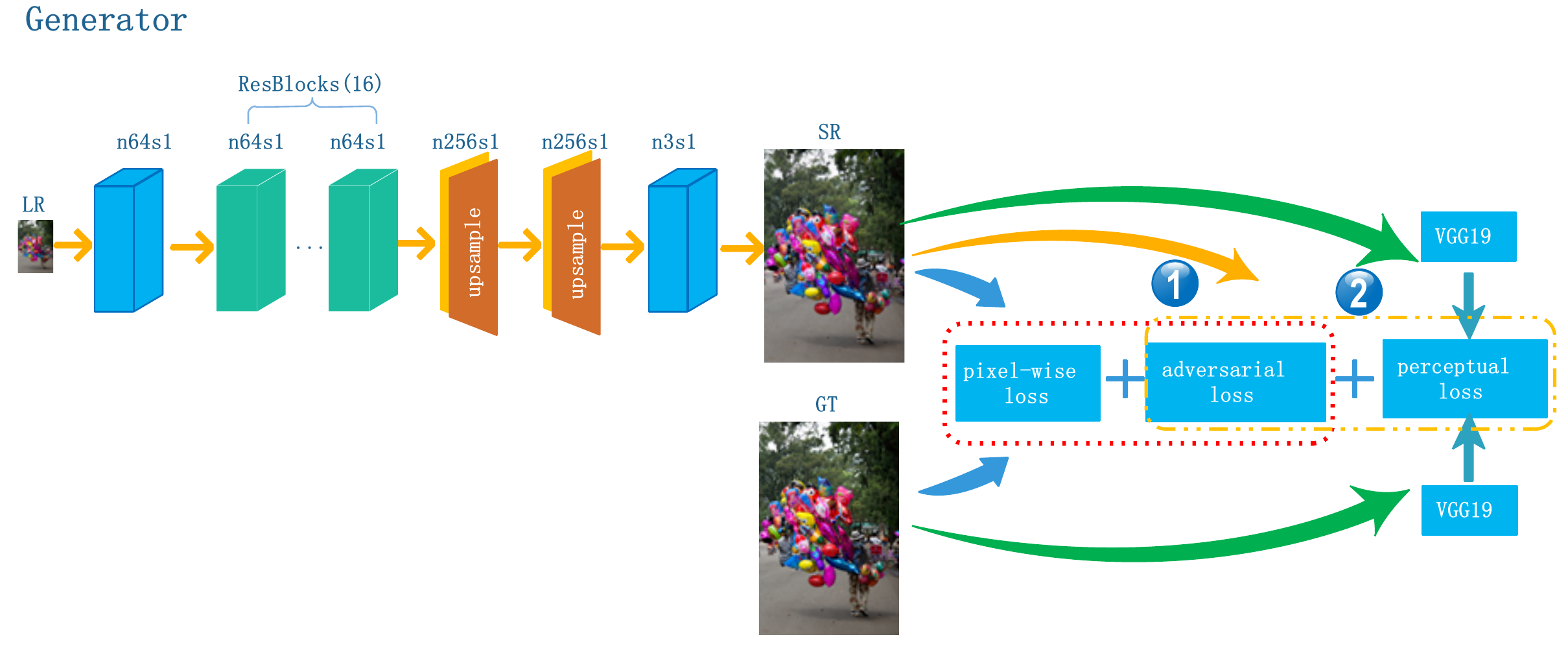}
\caption{The generator architecture of WP-SRGAN. The red box shows the generator loss in the first training stage, then the orange box contains the generator loss in the second training stage.}
\label{fig:3}
\end{figure}

{\bfseries Network architecture.} In WP-SRGAN, we use 16 ResBlocks in the generator network which is depicted in Fig. \ref{fig:3}. Each ResBlock is consisted of convolutional layer, PReLU activation layer and convolutional layer. And Batch Normalization (BN) layers are removed in both Generator and Discriminator. The architecture of Discriminator in WP-SRGAN is the same as MR-SRGAN except for removing BN layers.

 {\noindent\bfseries Loss function.} As shown in Fig. \ref{fig:3}, a two-stage bias adversarial training mechanism is adopted in WP-SRGAN by using different Generator losses. In the first stage, as the red box shows, we optimize the Generator loss which is consisted of pixel-wise loss and adversarial loss to obtain better objective performance (\emph{i.e.}, reduce the RMSE value). In the second stage, as the orange box shows, we regard the network parameters in the first stage as the pre-trained model and then replace the aforementioned generator loss with perceptual loss and adversarial loss to optimize for improving the subjective visual effects (\emph{e.g.}, reduce the perceptual index). The two-stage losses are represented as Eq. (\ref{eq6}) and Eq. (\ref{eq7}).
\begin{align}
  L_{1}=L_{pixel}+\lambda_1 L_{adv}\label{eq6}
\end{align}
\begin{align}
  L_{2}=\lambda_1 L_{adv}+\lambda_2 L_{vgg}\label{eq7}
\end{align}

Here, the pixel-wise loss is defined as the Eq. (\ref{eq3}), the perceptual loss adopts MSE loss by the features extracted from pre-trained VGG19 network, and the adversarial loss is donated as follows,
\begin{align}
  L_{adv}=-log(D(G(x_{l})))\label{eq8}
\end{align}

By adopting two-stage adversarial training mechanism, it can make the generated SR image similar to the corresponding ground truth in high-level features space.

\subsection{Soft-thresholding}

We can obtain different SR results by the two GANs aforementioned. One is MR-SRGAN, which emphasizes on improving the objective performance. The other is WP-SRGAN, which obtains the result that favors better subjective perception. To balance the perceptual score and RMSE of SR results, soft thresholding method proposed by Deng \emph{et al.} \cite{deng2018enhancing:27} is adopted to fuse the two SR images (\emph{i.e.}, MR-SRGAN, WP-SRGAN) which can be regarded as a way of pixel interpolation. The formulas are shown as follows,
\begin{align}
  I_{e}=I_{G}+soft(\Delta ,\xi  )\label{eq9}
\end{align}
\begin{align}
  soft(\Delta ,\xi  )=sign(\Delta )\cdot max(\left | \Delta  \right  |-\xi  ,0) \label{eq10}
\end{align}
where $I_{e}$ is the fused image, $\Delta =I_{G}-I_{g}$, $I_{G}$ is the generated image by WP-SRGAN whose perceptual score is lower, $I_{g}$ is the generated image by MR-SRGAN whose RMSE value is lower. $\xi$ is the adjusted threshold which is discussed in Section 4.2.

\section{Experimental Results}

In this section, we conduct extensive experiments on five publicly available benchmarks for scaling factor $4\times$ image SR: Set5 \cite{bevilacqua2012low:28}, Set14 \cite{zeyde2010single:29}, B100 \cite{arbelaez2011contour:30}, Urban100 \cite{huang2015single:31}, Managa109 \cite{matsui2017sketch:32}, separately. The first three datasets Set5, Set14, BSD100 mainly contain natural images, Urban100 consists of 100 urban images, and Manga109 is Japanese anime containing fewer texture features. Then we compare the performance of our proposed Bi-GANs-ST algorithm with the state-of-the-art SR algorithms in terms of objective criteria and subjective visual perception.

\subsection{Implementation and training details}
We train our networks using the RAISE\footnote{http://loki.disi.unitn.it/RAISE/}  dataset which consists of $8156$ HR RAW images. The HR images are downsampling by bicubic interpolation method for the scaling factor $4\times$ to obtain the LR images. To analyze our models capacity, we evaluate them on the PIRM-SR $2018$ self validation dataset \cite{Blau2018real:12} which consists of $100$ realistic images including human, plants, animals and so on.

The LR-HR image patches for training are randomly cropped from the corresponding LR and HR image pairs. The crop size for LR patches is $24\times24$, and the size of corresponding HR patches is $96\times96$. Random flipping is used for image argumentation. The batch size is set to $16$.

In our experiments, MP-SRGAN is conducted on the deep learning framework, \emph{i.e.}, Pytorch and WP-SRGAN is conducted on Tensorflow. We implement our method on the platform Ubuntu 16.04, CUDA8.0 and CUDNN6.0 with GTX1080 GPU and 32G CPU Memory.

In Generator of MR-SRGAN, $7$ MR blocks are used. The filter size is set to $3\times3$. The learning rate is initialized to $1e-4$ and Adam optimizer with the momentum $0.9$ is utilized. The network is trained for $600$ epochs, and we choose the best results according to the metric SSIM.

In Generator of WP-SRGAN, 16 ResBlocks are used and the filter size is $3\times3$. The filter size is $9\times9$ in the first and last convolutional layer. All the convolutional layers use one stride and one padding. The weights are initialized by Xavier method. All the convolutional and upsampling layers are followed by PReLU activation function. The learning rate is initialized to $1e-4$ and decreased by a factor of $10$ for $2.5\times10^{5}$ iterations and total iterations are $5\times10^{5}$. We use Adam optimizer with momentum $0.9$. In Discriminator, the filter size is $3\times3$, and the number of features is twice increased from $64$ to $512$, the stride is one or two, alternately.

The weights of adversarial loss and perceptual loss both in MP-SRGAN and WP-SRGAN (\emph{i.e.}, $\lambda_1$ and $\lambda_2$) are set to $1e-3$, $6e-3$, respectively. And the threshold (\emph{i.e.}, $\xi$) for image fusion is set to 0.73 in our experiment.

\setlength{\tabcolsep}{4pt}
\begin{table}[h]
\begin{center}
\caption{The quantitative results of WP-SRGAN with two stages on the PRIM 2018 self validation dataset for $4\times$ enlargement.}
\label{table:1}
\begin{tabular}{ll}
\hline\noalign{\smallskip}
Model & Perceptual score/RMSE)\\
\noalign{\smallskip}
\hline
\noalign{\smallskip}
First stage & 5.2002 / 14.385\\
Second stage & 2.0815 / 16.2813\\
\hline
\end{tabular}
\end{center}
\end{table}

\subsection{Model analysis}
\begin{figure}[t]
\centering
\setlength{\abovecaptionskip}{0.cm}
\setlength{\belowcaptionskip}{-0.cm}
\centering
\begin{tabular}{p{3.3cm}<{\centering}  p{3.3cm}<{\centering} p{3.3cm}<{\centering}} HR & First stage & Second stage \end{tabular}\\
 \begin{overpic}[scale=.3]{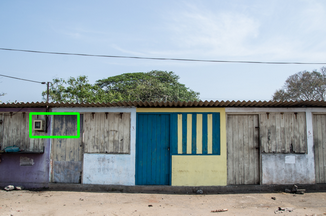} \put(50,42){\includegraphics[scale=0.5]{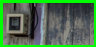}} \end{overpic} \
 \begin{overpic}[scale=.3]{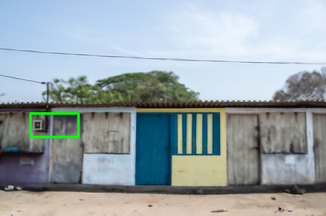} \put(50,42){\includegraphics[scale=0.5]{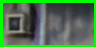}} \end{overpic} \
 \begin{overpic}[scale=.3]{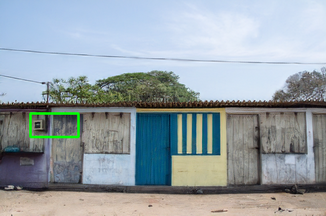} \put(50,42){\includegraphics[scale=0.5]{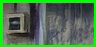}} \end{overpic} \vspace{-0mm}
 \\
  \begin{overpic}[scale=.3]{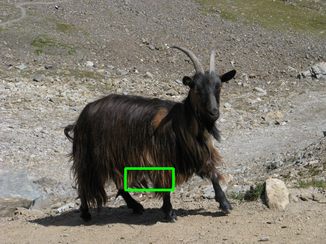} \put(50,53){\includegraphics[scale=0.5]{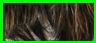}} \end{overpic} \
 \begin{overpic}[scale=.3]{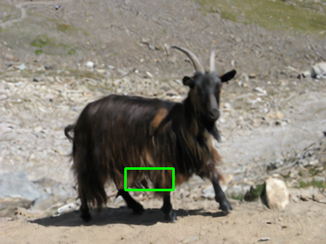} \put(50,53){\includegraphics[scale=0.5]{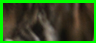}} \end{overpic} \
 \begin{overpic}[scale=.3]{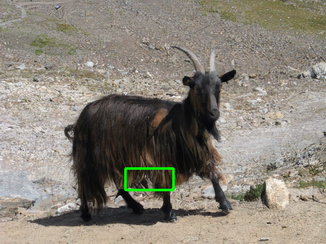} \put(50,53){\includegraphics[scale=0.5]{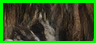}} \end{overpic} \vspace{-0mm}

\caption{The visual results of WP-SRGAN with two stages on the PRIM 2018 self validation dataset for $4\times$ enlargement.}
\label{fig:4}
\end{figure}

{\bfseries Training WP-SRGAN with Two stages.}  We analyze the experimental results of the two-stage adversarial training mechanism in WP-SRGAN. The quantitative and qualitative results on PIRM-SR $2018$ self validation dataset are shown in Table \ref{table:1} and Fig. \ref{fig:4}.

In Table \ref{table:1}, WP-SRGAN with two stages can achieve lower perceptual score than WP-SRGAN with the first stage. As shown in Fig. \ref{fig:4}, the recovered details of WP-SRGAN with two stages are much more than WP-SRGAN with the first stage. And the images generated by two stages look more realistic. Therefore, we use WP-SRGAN with two stages in our model.

\begin{figure}[h]
\centering
\includegraphics[scale=0.5]{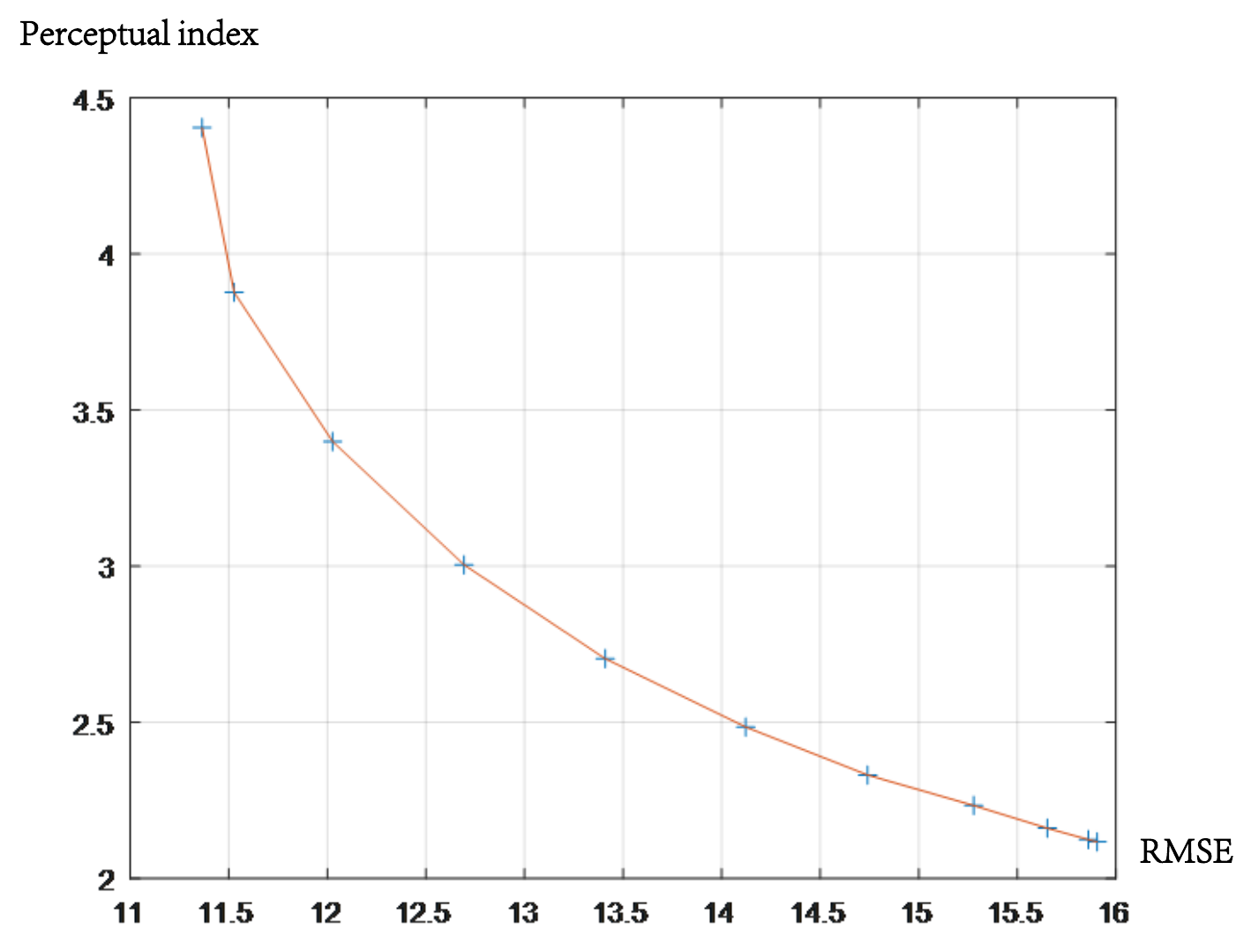}
\caption{The perceptual-distortion plane of our method. The points on the curve denote the different thresholds from $0$ to $1$ with an interval of $0.1$.}
\label{fig:5}
\end{figure}

{\noindent\bfseries Soft thresholding.} In the challenge, three regions are defined by RMSE between $11.5$ and $16$. According to different threshold settings, we draw the perceptual-distoration plane which is shown in Fig. \ref{fig:5}, according to the results fused by Eq. (\ref{eq9}) and (\ref{eq10}). The points on the curve denote the different thresholds from $0$ to $1$ with an interval of $0.1$. Experimental results show that we can obtain excellent perceptual score in Region3 (RMSE is between $12.5$ and $16$) when $\xi$ is set to 0.73.

\begin{figure}[t]
\centering
\setlength{\abovecaptionskip}{0.cm}
\setlength{\belowcaptionskip}{-0.cm}
\begin{tabular}{p{2.8cm}<{\centering}  p{2.8cm}<{\centering} p{2.8cm}<{\centering} p{2.8cm}<{\centering}} HR & MR-SRGAN & WP-SRGAN & Bi-GANs-ST \end{tabular}\\
 \begin{overpic}[scale=.23]{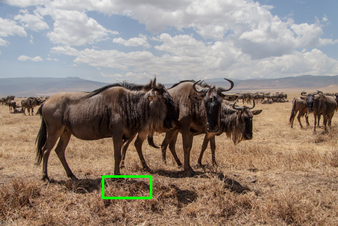} \put(40,40){\includegraphics[scale=0.48]{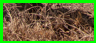}} \end{overpic} \
 \begin{overpic}[scale=.23]{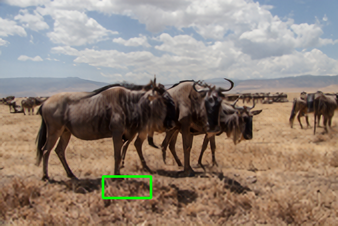} \put(40,40){\includegraphics[scale=0.48]{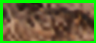}} \end{overpic} \
 \begin{overpic}[scale=.23]{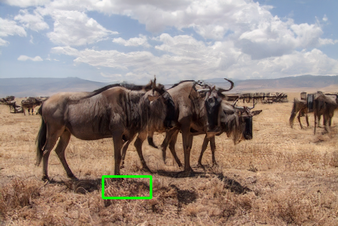} \put(40,40){\includegraphics[scale=0.48]{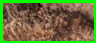}} \end{overpic} \
 \begin{overpic}[scale=.23]{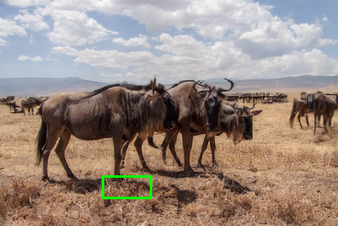} \put(40,40){\includegraphics[scale=0.48]{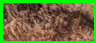}} \end{overpic} \vspace{-0mm}
 \\
 \begin{overpic}[scale=.24]{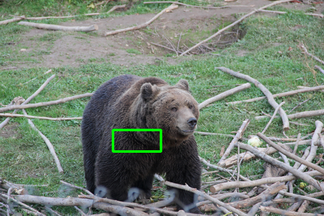} \put(40,40){\includegraphics[scale=0.48]{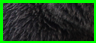}} \end{overpic} \
 \begin{overpic}[scale=.24]{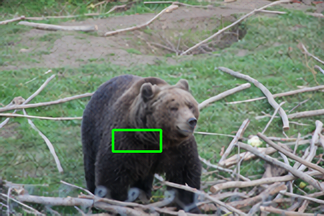} \put(40,40){\includegraphics[scale=0.48]{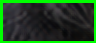}} \end{overpic} \
 \begin{overpic}[scale=.24]{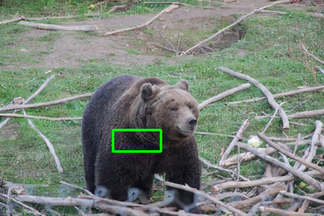} \put(40,40){\includegraphics[scale=0.48]{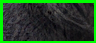}} \end{overpic} \
 \begin{overpic}[scale=.24]{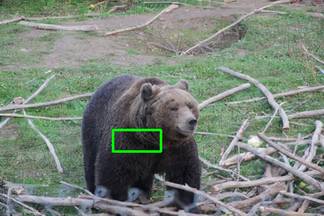} \put(40,40){\includegraphics[scale=0.48]{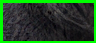}} \end{overpic} \vspace{-0mm}
\caption{The visual results on three models (MR-SRGAN, WP-SRGAN and Bi-GANs-ST) for scaling factor $4\times$ for the metrics perceptual score and RMSE on the PIRM-SR 2018 self validation dataset.}
\label{fig:6}
\end{figure}

{\noindent\bfseries Model capacity.} To demonstrate the capability of our models, we analyze the SR results of MR-SRGAN, WP-SRGAN and Bi-GANs-ST for the metrics perceptual score and RMSE on the PIRM-SR 2018 self validation dataset. The quantitative and qualitative results are shown in Table \ref{table:2} and Fig. \ref{fig:6}. The experimental results show that Bi-GANs-ST can keep balance between the perceptual score and RMSE.

\setlength{\tabcolsep}{1.4pt}
\setlength{\tabcolsep}{4pt}
\begin{table}[h]
\begin{center}
\caption{The model capacity analysis of the SR results by MR-SRGAN, WP-SRGAN and Bi-GANs-ST for the metrics perceptual score and RMSE on the PIRM-SR 2018 self validation dataset.}
\label{table:2}
\begin{tabular}{ll}
\hline\noalign{\smallskip}
Model & Perceptual score/RMSE\\
\noalign{\smallskip}
\hline
\noalign{\smallskip}
MR-SRGAN & 4.404 / 11.36\\
WP-SRGAN & 2.082 / 16.28\\
Bi-GANs-ST & 2.139/ 15.77\\
\hline
\end{tabular}
\end{center}
\end{table}
\setlength{\tabcolsep}{1.4pt}

\begin{figure}
\centering
\setlength{\abovecaptionskip}{0.cm}
\setlength{\belowcaptionskip}{-0.cm}
\tiny
\begin{tabular}{p{1.5cm}<{\centering} p{1.5cm}<{\centering} p{1.5cm}<{\centering} p{1.5cm}<{\centering} p{1.5cm}<{\centering} p{1.5cm}<{\centering} p{1.5cm}<{\centering}} Bicubic & EDSR \cite{lim2017enhanced:6} & EnhanceNet \cite{sajjadi2017enhancenet:8} & MR-SRGAN (ours) & WP-SRGAN (ours) &  Bi-GANs-ST (ours) & Ground Truth  \end{tabular}\\
\includegraphics[width=.13\textwidth]{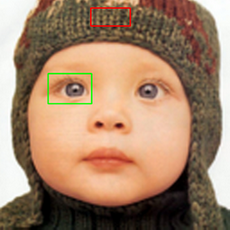}
\includegraphics[width=.13\textwidth]{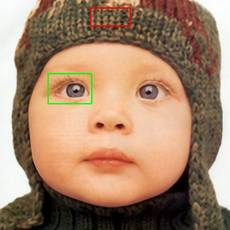}
\includegraphics[width=.13\textwidth]{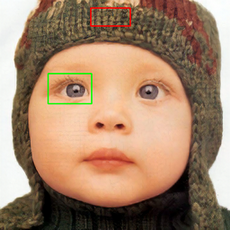}
\includegraphics[width=.13\textwidth]{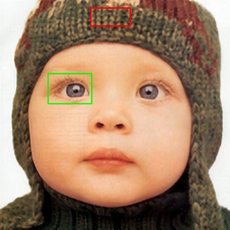}
\includegraphics[width=.13\textwidth]{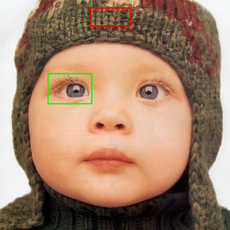}
\includegraphics[width=.13\textwidth]{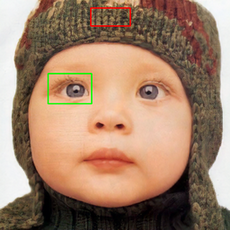}
\includegraphics[width=.13\textwidth]{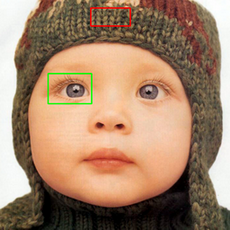}
\\
\includegraphics[width=.13\textwidth]{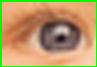}
\includegraphics[width=.13\textwidth]{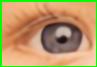}
\includegraphics[width=.13\textwidth]{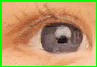}
\includegraphics[width=.13\textwidth]{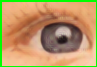}
\includegraphics[width=.13\textwidth]{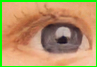}
\includegraphics[width=.13\textwidth]{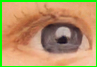}
\includegraphics[width=.13\textwidth]{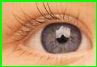}
\\
\includegraphics[width=.13\textwidth]{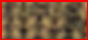}
\includegraphics[width=.13\textwidth]{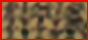}
\includegraphics[width=.13\textwidth]{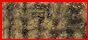}
\includegraphics[width=.13\textwidth]{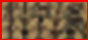}
\includegraphics[width=.13\textwidth]{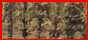}
\includegraphics[width=.13\textwidth]{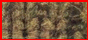}
\includegraphics[width=.13\textwidth]{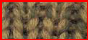}
\\
\includegraphics[width=.13\textwidth]{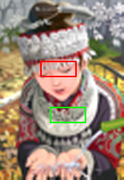}
\includegraphics[width=.13\textwidth]{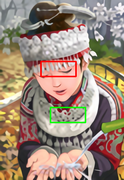}
\includegraphics[width=.13\textwidth]{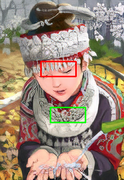}
\includegraphics[width=.13\textwidth]{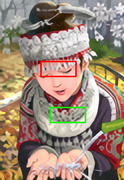}
\includegraphics[width=.13\textwidth]{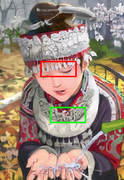}
\includegraphics[width=.13\textwidth]{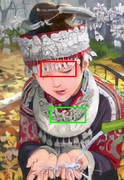}
\includegraphics[width=.13\textwidth]{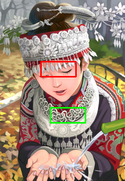}
\\
\includegraphics[width=.13\textwidth]{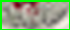}
\includegraphics[width=.13\textwidth]{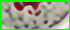}
\includegraphics[width=.13\textwidth]{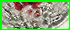}
\includegraphics[width=.13\textwidth]{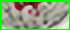}
\includegraphics[width=.13\textwidth]{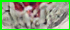}
\includegraphics[width=.13\textwidth]{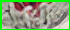}
\includegraphics[width=.13\textwidth]{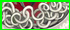}
\\
\includegraphics[width=.13\textwidth]{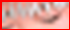}
\includegraphics[width=.13\textwidth]{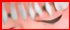}
\includegraphics[width=.13\textwidth]{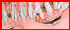}
\includegraphics[width=.13\textwidth]{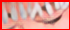}
\includegraphics[width=.13\textwidth]{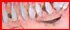}
\includegraphics[width=.13\textwidth]{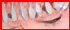}
\includegraphics[width=.13\textwidth]{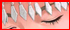}
\\
\includegraphics[width=.13\textwidth]{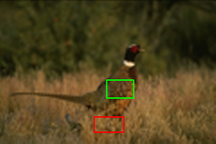}
\includegraphics[width=.13\textwidth]{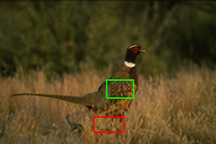}
\includegraphics[width=.13\textwidth]{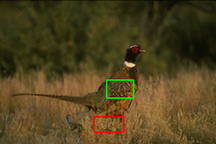}
\includegraphics[width=.13\textwidth]{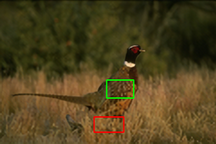}
\includegraphics[width=.13\textwidth]{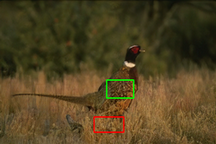}
\includegraphics[width=.13\textwidth]{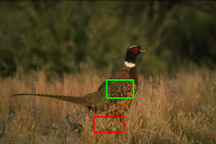}
\includegraphics[width=.13\textwidth]{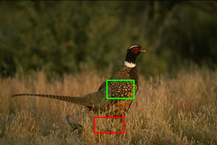}
\\
\includegraphics[width=.13\textwidth]{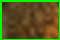}
\includegraphics[width=.13\textwidth]{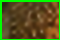}
\includegraphics[width=.13\textwidth]{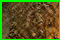}
\includegraphics[width=.13\textwidth]{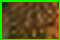}
\includegraphics[width=.13\textwidth]{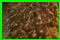}
\includegraphics[width=.13\textwidth]{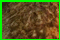}
\includegraphics[width=.13\textwidth]{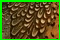}
\\
\includegraphics[width=.13\textwidth]{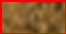}
\includegraphics[width=.13\textwidth]{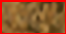}
\includegraphics[width=.13\textwidth]{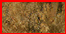}
\includegraphics[width=.13\textwidth]{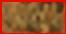}
\includegraphics[width=.13\textwidth]{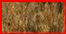}
\includegraphics[width=.13\textwidth]{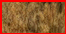}
\includegraphics[width=.13\textwidth]{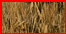}
\\
\includegraphics[width=.13\textwidth]{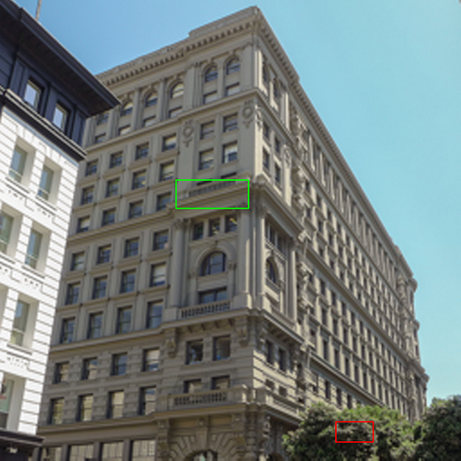}
\includegraphics[width=.13\textwidth]{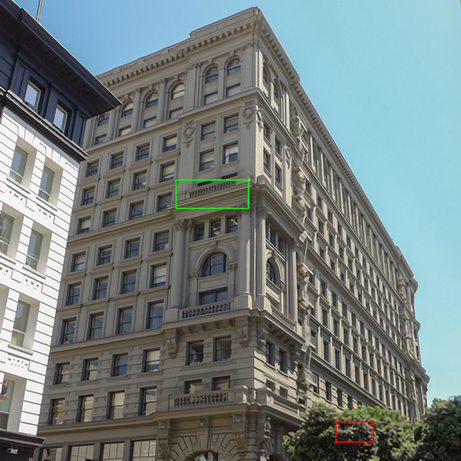}
\includegraphics[width=.13\textwidth]{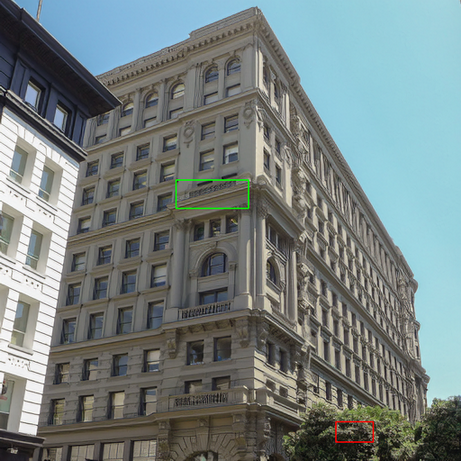}
\includegraphics[width=.13\textwidth]{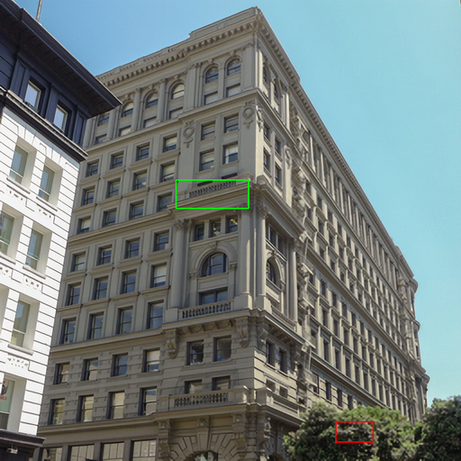}
\includegraphics[width=.13\textwidth]{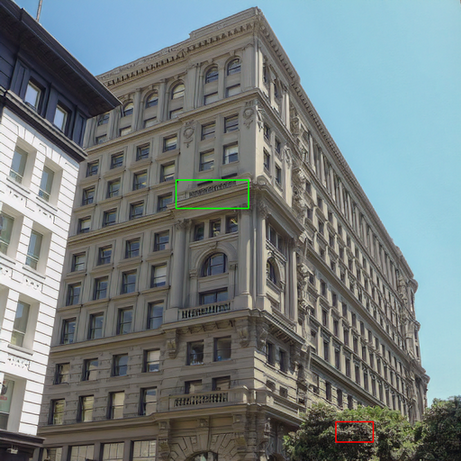}
\includegraphics[width=.13\textwidth]{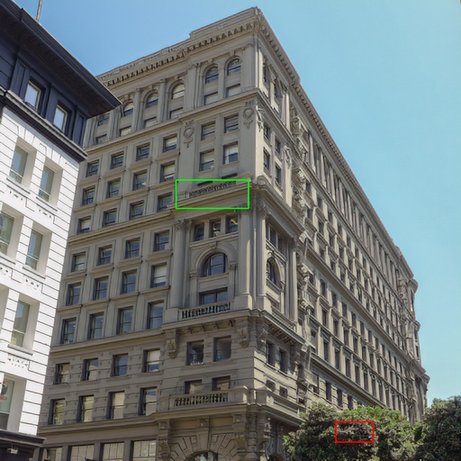}
\includegraphics[width=.13\textwidth]{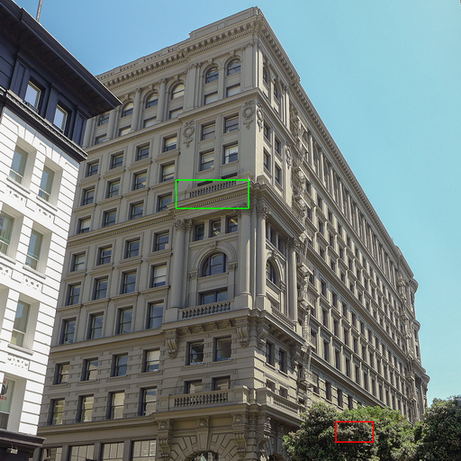}
\\
\includegraphics[width=.13\textwidth]{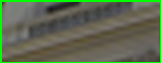}
\includegraphics[width=.13\textwidth]{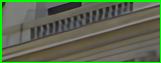}
\includegraphics[width=.13\textwidth]{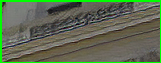}
\includegraphics[width=.13\textwidth]{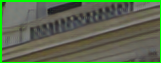}
\includegraphics[width=.13\textwidth]{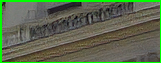}
\includegraphics[width=.13\textwidth]{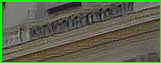}
\includegraphics[width=.13\textwidth]{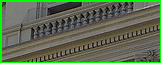}
\\
\includegraphics[width=.13\textwidth]{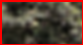}
\includegraphics[width=.13\textwidth]{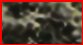}
\includegraphics[width=.13\textwidth]{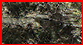}
\includegraphics[width=.13\textwidth]{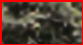}
\includegraphics[width=.13\textwidth]{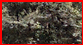}
\includegraphics[width=.13\textwidth]{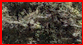}
\includegraphics[width=.13\textwidth]{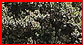}
\\
\includegraphics[width=.13\textwidth]{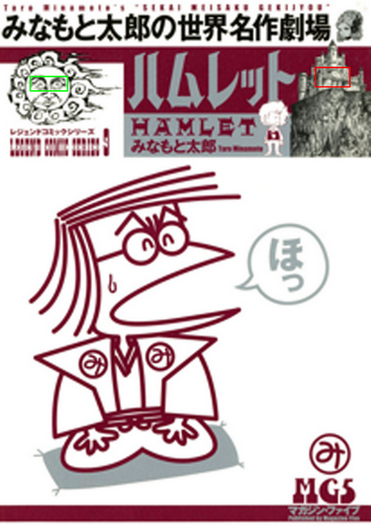}
\includegraphics[width=.13\textwidth]{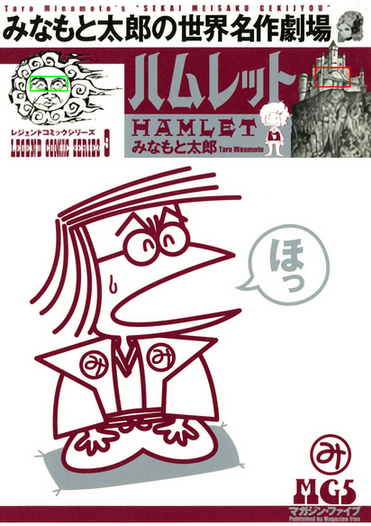}
\includegraphics[width=.13\textwidth]{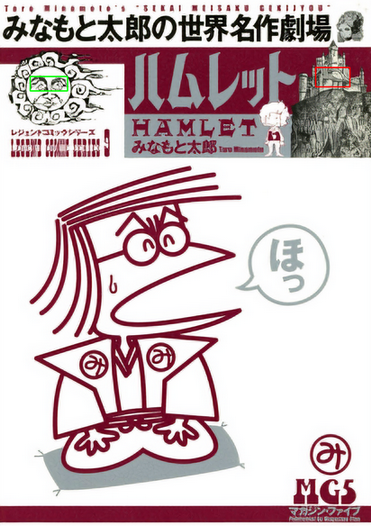}
\includegraphics[width=.13\textwidth]{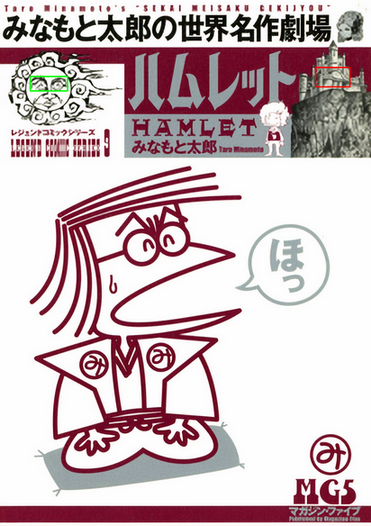}
\includegraphics[width=.13\textwidth]{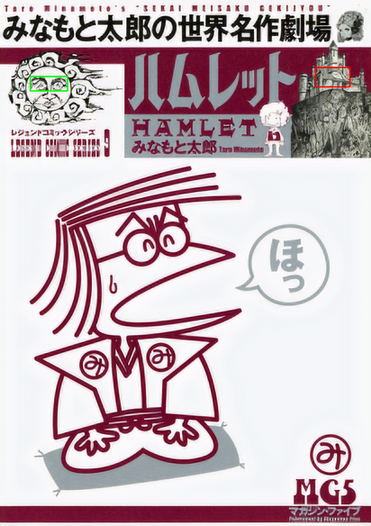}
\includegraphics[width=.13\textwidth]{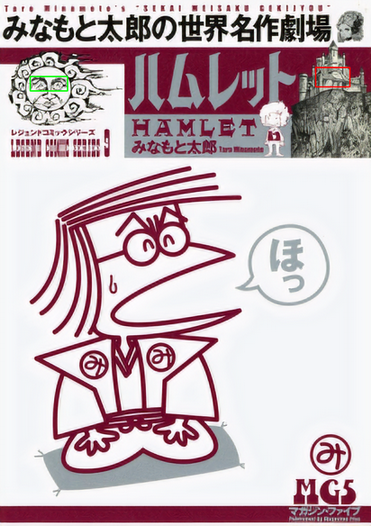}
\includegraphics[width=.13\textwidth]{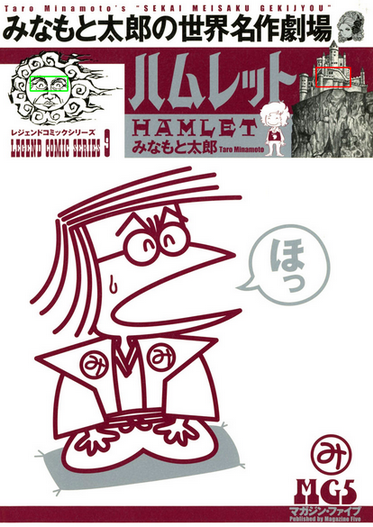}
\\
\includegraphics[width=.13\textwidth]{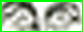}
\includegraphics[width=.13\textwidth]{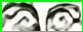}
\includegraphics[width=.13\textwidth]{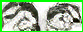}
\includegraphics[width=.13\textwidth]{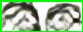}
\includegraphics[width=.13\textwidth]{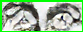}
\includegraphics[width=.13\textwidth]{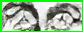}
\includegraphics[width=.13\textwidth]{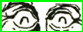}
\\
\includegraphics[width=.13\textwidth]{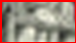}
\includegraphics[width=.13\textwidth]{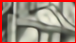}
\includegraphics[width=.13\textwidth]{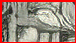}
\includegraphics[width=.13\textwidth]{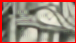}
\includegraphics[width=.13\textwidth]{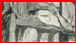}
\includegraphics[width=.13\textwidth]{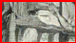}
\includegraphics[width=.13\textwidth]{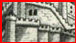}
\\
\caption{The visual results on five benchmark datasets for scaling factor $4\times$ which is Bicubic, EDSR, EnhanceNet, MR-SRGAN, WP-SRGAN, Bi-GANs-ST, ground truth from left to right.}
\label{fig:7}
\end{figure}
\subsection{Comparison with the state-of-the-arts}
To verificate the validity of our Bi-GANs-ST, we conduct extensive experiments on five publicly available benchmarks and compare the results with other state-of-the-art SR algorithms, including EDSR\cite{lim2017enhanced:6}, EnhanceNet\cite{sajjadi2017enhancenet:8}. We use the open-source implementations for the two comparison methods. We evaluate the SR images with image quality assessment indices (\emph{i.e.}, PSNR, SSIM, perceptual score, RMSE) where PSNR and SSIM are measured on the y channel and ignored 6 pixels from the border.

The quantitative results for evaluating PSNR and SSIM are shown in Table \ref{table:3}. The best algorithm is EDSR, which is on average $1.0dB$, $0.54dB$, $0.34dB$, $0.83dB$ and $1.13dB$ higher than our MR-SRGAN. The PSNR values of our Bi-GANs-ST are higher than EnhanceNet on Set5, Urban100, Manga109 approximately $0.64dB$, $0.3dB$, $0.13dB$, respectively. The SSIM values of our Bi-GANs-ST are all higher than EnhanceNet. Table \ref{table:4} shows the quantitative evaluation of average perceptual score and RMSE. For perceptual score index, our WP-SRGAN achieves the best and Bi-GANs-ST achieves the second best on five benchmarks except for Set5. For RMSE index, EDSR performs the best and our MR-SRGAN performs the second best.

\setlength{\tabcolsep}{4pt}
\begin{table}
\scriptsize
\begin{center}
\caption{Quantitative evaluation of state-of-the-art SR algorithms on five publicly available benchmarks: average PSNR/SSIM for scaling factor $4\times$ (Red text indicates the best and blue text indicates the second best performance).}
\label{table:3}
\begin{tabular}{llllll}
\hline\noalign{\smallskip}
Algorithm & Set5 & Set14 & BSD100 & Urban100 & Manga109\\
\noalign{\smallskip}
\hline
\noalign{\smallskip}
Bicubic & 24.74/0.736 & 23.47/0.630 & 23.93/0.602 & 21.15/0.583 & 21.82/0.711\\
EDSR\cite{lim2017enhanced:6} & {\color{red}32.53/0.899} & {\color{red}28.82/0.786} & {\color{red}27.64/0.740} & {\color{red}26.62/0.802} & {\color{red}30.95/0.914}\\
MR-SRGAN(ours) & {\color{blue}31.53/0.884} & {\color{blue}28.28/0.771} & {\color{blue}27.30/0.725} & {\color{blue}25.79/0.774} & {\color{blue}29.82/0.895}\\
\hline
EnhanceNet\cite{sajjadi2017enhancenet:8} & 28.90/0.818 & 26.04/0.685 & 25.19/0.634 & 23.60/0.691 & 26.71/0.827\\
WP-SRGAN(ours) & 29.06/0.834 & 26.01/0.703 & 24.38/0.642 & 23.74/0.700 & 	26.62/0.836\\
Bi-GANs-ST(ours) & 29.54/0.840 & 26.01/0.706 & 24.54/0.651 & 23.90/0.703 & 	26.84/0.839\\
\hline
\end{tabular}
\end{center}
\end{table}
\setlength{\tabcolsep}{1.4pt}

\setlength{\tabcolsep}{4pt}
\begin{table}
\scriptsize
\begin{center}
\caption{Quantitative evaluation of state-of-the-art SR algorithms on five publicly available benchmarks: average perceptual scores/RMSE for scale $4\times$ (Red text indicates the best and blue text indicates the second best performance).}
\label{table:4}
\begin{tabular}{llllll}
\hline\noalign{\smallskip}
Algorithm & Set5 & Set14 & BSD100 & Urban100 & Manga109\\
\noalign{\smallskip}
\hline
\noalign{\smallskip}
EDSR\cite{lim2017enhanced:6} & 5.944/{\color{red}6.51} & 5.256/{\color{red}10.92} & 5.263/{\color{red}12.38} & 4.981/{\color{red}14.52} & 4.714/{\color{red}8.63}\\
EnhanceNet\cite{sajjadi2017enhancenet:8} & {\color{red}3.141}/9.94 & 3.009/14.82 & 2.979/16.14 & 3.401/19.49 & 3.259/13.19\\
MR-SRGAN(ours) & 5.304/{\color{blue}7.44} &4.703/{\color{blue}11.41} & 4.995/{\color{blue}12.72} & 4.466/{\color{blue}15.64} & 4.164/{\color{blue}9.57}\\
WP-SRGAN(ours) & {\color{blue}3.317}/9.80 & {\color{red}2.824}/14.46 & {\color{red}2.226}/18.19 & {\color{red}3.260}/19.04 & {\color{red}3.195}/13.32\\
Bi-GANs-ST(ours) & 3.531/9.14 & {\color{blue}2.869}/14.22 & {\color{blue}2.375}/17.56 & {\color{blue}3.272}/18.75 & {\color{blue}3.216}/12.97\\
\hline
\end{tabular}
\end{center}
\end{table}
\setlength{\tabcolsep}{1.4pt}

The visual perception results of $4\times$ enlargement of different algorithms on five benchmarks are shown in Fig. \ref{fig:7}. These visual results are produced by Bicubic, EDSR, EnhanceNet, MR-SRGAN, WP-SRGAN, Bi-GANs-ST and the ground truth from left to right. EDSR can generate the images which look clear and smooth but not realistic. The SR images of our MR-SRGAN algorithm are like to EDSR. EnhanceNet can generate more realistic images with unpleasant noises. The SR images of our WP-SRGAN algorithm obtain more details like EnhanceNet with less noises which are more close to the ground-truth. And our Bi-GANs-ST algorithm has fewer noises than WP-SRGAN.

\section{Conclusions}
In this paper, we propose a new deep SR framework Bi-GANs-ST by integrating two complementary generative adversarial networks (GAN) branches. To keep better balance between the perceptual score and RMSE of generated images, we redesign two GANs (\emph{i.e.}, MR-SRGAN, WP-SRGAN) to generate two complementary SR results based on SRGAN. Last, we use soft-thresholding method to fuse two SR results which can make the perceptual score and RMSE tradeoff. Experimental results on five publicly benchmarks show that our proposed algorithm can perform better perceptual results than other SR algorithms for $4\times$ enlargement.\\

{\noindent\bfseries Acknowledgements.} This work is supported by the National Natural Science Foundation of China under Grant 61876161, Grant 61772524, Grant 61373077 and in part by the Beijing Natural Science Foundation under Grant 4182067.

\bibliographystyle{splncs}
\bibliography{egbib}
\end{document}